\documentclass[conference,10pt]{IEEEtran}

\usepackage{makecell}
\usepackage{cite}
\usepackage{url}
\usepackage{amsmath,amssymb,amsfonts}
\usepackage{verbatim}
\usepackage{dblfloatfix}
\usepackage{algorithm}
\usepackage{algpseudocode}
\usepackage{multirow}
\usepackage{booktabs}
\usepackage{graphicx}
\usepackage{caption}
\usepackage{subcaption}
\usepackage[table]{xcolor}
\usepackage{hyperref}

\makeatletter
\g@addto@macro\normalsize{%
  \setlength{\abovedisplayskip}{1pt}%
  \setlength{\belowdisplayskip}{1pt}%
  \setlength{\abovedisplayshortskip}{1pt}%
  \setlength{\belowdisplayshortskip}{1pt}%
}
\makeatother

\IEEEoverridecommandlockouts
\IEEEpubid{%
  \makebox[\columnwidth]{%
    \parbox{\columnwidth}{\scriptsize © 2026 IEEE. Personal use of this material is permitted. Permission from IEEE must be obtained for all other uses.\hfill}%
  }%
  \hspace{\columnsep}%
  \makebox[\columnwidth]{}%
}

\begin{document}
\bstctlcite{IEEEexample:BSTcontrol}

\title{Meta-Learning for Data-Efficient Plant Growth Estimation via Vision Transformers and Fuzzy Clustering\\
\thanks{This research was funded by the Akershus Fylkeskommune (project number 2024-0656) and Photosynthetic AS.}}

\author{
\IEEEauthorblockN{
Sheikh Hasan Elahi\IEEEauthorrefmark{1},
Rusith Chamara Hathurusinghe Dewage\IEEEauthorrefmark{1}\IEEEauthorrefmark{3},
Habib Ullah\IEEEauthorrefmark{1},\\
Muhammad Salman Siddiqui\IEEEauthorrefmark{2},
Rakibul Islam\IEEEauthorrefmark{3},
Fadi Al Machot\IEEEauthorrefmark{1}
}
\IEEEauthorblockA{\IEEEauthorrefmark{1}Department of Data Science, Norwegian University of Life Sciences (NMBU), Norway}
\IEEEauthorblockA{\IEEEauthorrefmark{2}Department of Mechanical Engineering and Technology Management, Norwegian University of Life Sciences (NMBU), Norway}
\IEEEauthorblockA{\IEEEauthorrefmark{3}Photosynthetic AS, Hegdehaugsveien 24, 0352 Oslo, Norway}
\IEEEauthorblockA{Email: \{sheikh.hasan.elahi, rusith.chamara.hathurusinghe.dewage, habib.ullah, muhammad.salman.siddiqui, fadi.al.machot\}@nmbu.no}
\IEEEauthorblockA{Email: rakibul@photosynthetic.com}
}

\maketitle

\IEEEpubidadjcol

\begin{abstract}
Accurate plant growth estimation is essential for greenhouse monitoring, yet obtaining labeled data remains costly and time-consuming. To address this, we propose a few-shot regression framework that combines Vision Transformer (ViT) feature embeddings, clustering-based task construction, and gradient-based meta-learning, and show that task construction in embedding space is a primary driver of performance. The approach leverages an unlabeled image pool to organize data into structured tasks using fuzzy c-means clustering, enabling efficient learning from a small number of labeled samples.
We systematically evaluate meta-learning methods and show that second-order methods (e.g., Model-Agnostic Meta-Learning variants such as MAML++) outperform classical baselines in the few-shot regime. Furthermore, intra-cluster support selection has a limited and dataset-dependent impact.
Experiments on two plant datasets show that structured task design combined with meta-learning enables reliable plant growth estimation under severe label scarcity.
\end{abstract}

\section{Introduction}
Accurate estimation of plant growth is essential for controlled-environment agriculture, where timely monitoring enables optimized irrigation, fertigation, and yield prediction. Traditional phenotyping methods rely on manual measurements of traits such as plant height, leaf area, and canopy width. Although reliable, these approaches are labor-intensive and unsuitable for high-frequency monitoring in modern greenhouse and vertical farming systems \cite{singh2016}. Deep learning has significantly advanced automated phenotyping; however, most methods require large labeled datasets, which are difficult to obtain in controlled environments due to limited annotation capacity and variability \cite{alhnaity2020,singh2018}.

Few-shot learning (FSL) addresses this limitation by enabling models to generalize from a small number of labeled samples. Classical approaches such as Siamese networks \cite{koch2015} and prototypical networks \cite{snell2017} have demonstrated strong performance in low-data settings, with growing applications in agriculture \cite{yang2022,lagergren2023}.
Furthermore, meta-learning improves sample efficiency by learning model initializations that enable rapid adaptation to new tasks. The Model-Agnostic Meta-Learning (MAML) algorithm \cite{finn2017model} and its variants---Meta-SGD \cite{liMetaSGD2017} and MAML++ \cite{antoniou2019mamlpp}---have shown that meta-initializations can significantly improve performance in data-limited settings \cite{satrya2023}. Agricultural extensions such as SWE-MAML \cite{li2025} and potato growth estimation via meta-learning \cite{yang2025} further demonstrate the potential of this approach in phenotyping. Yet, a systematic comparison of MAML variants for image-based growth regression has not been performed.
In parallel, Vision Transformers (ViTs) \cite{dosovitskiy2020} provide powerful representations of plant morphology, capturing global structural information. When combined with clustering techniques such as fuzzy c-means (FCM) \cite{bezdek1981}, these embeddings enable structured sampling of representative examples, aligning with recent advances in coreset selection \cite{lee2024}. However, the integration of representation learning, clustering-based sampling, and meta-learning for plant growth estimation has not been thoroughly investigated.

In this work, we propose a unified framework for few-shot plant growth regression that combines ViT embeddings, FCM-based representative sampling, and gradient-based meta-learning. The framework is evaluated on two distinct datasets (cucumber and lettuce) to assess cross-dataset generalization. Our main contributions are:
(a) a clustering-based task construction strategy that organizes few-shot regression tasks in embedding space;
(b) a systematic analysis showing task construction dominates performance over sampling heuristics; and
(c) an empirical study demonstrating the importance of second-order meta-learning for plant growth regression.

The proposed approach demonstrates that meta-learning combined with structured sampling provides a practical solution for greenhouse phenotyping systems.
\section{Related Work}

\begin{figure*}[t]
\centering
\includegraphics[width=0.6\textwidth]{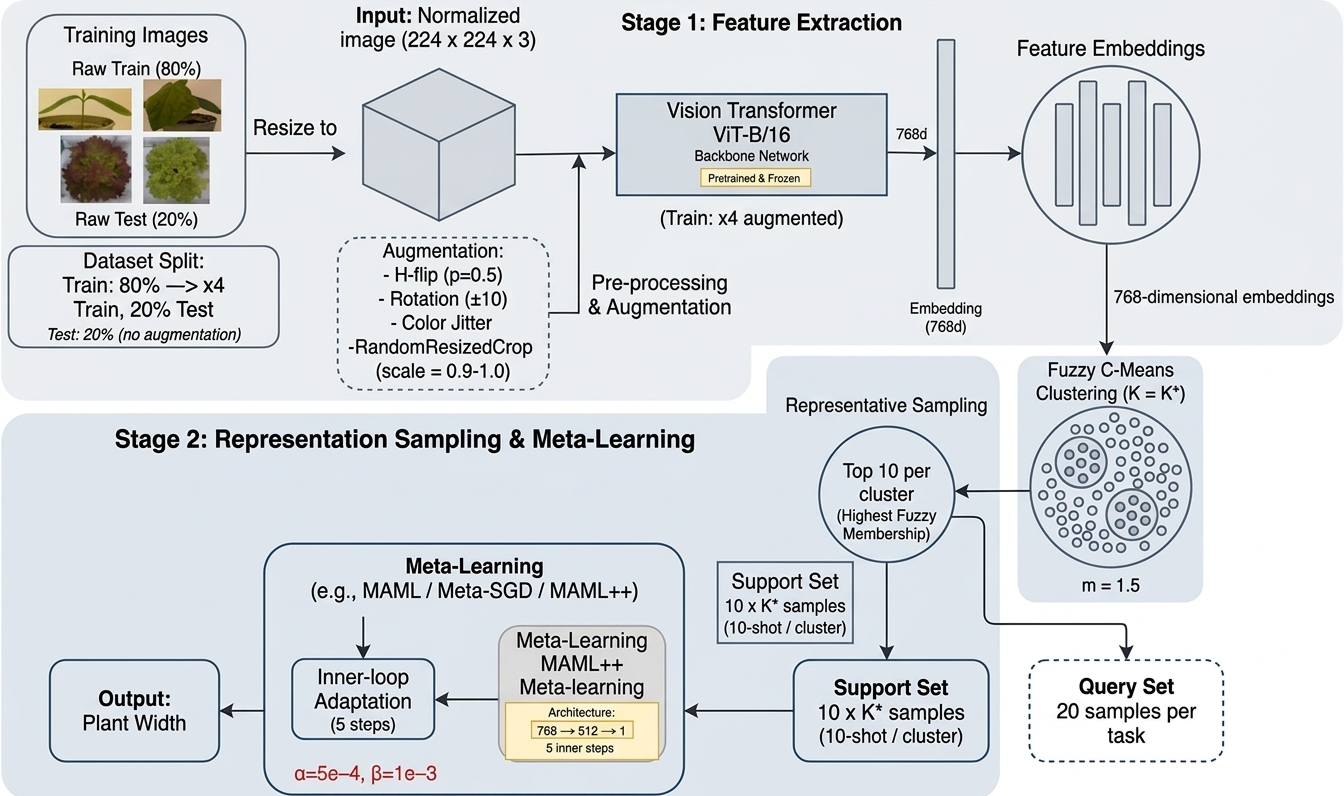}
\caption{Overview of the proposed pipeline. Images are encoded using a pretrained Vision Transformer, clustered via fuzzy c-means to capture morphological structure, and used to construct few-shot tasks for meta-learning. Each task corresponds to a local region in embedding space, enabling efficient adaptation for plant growth regression.}
\label{fig:pipeline}
\end{figure*}

\subsection{Few-Shot Learning in Agriculture}

Few-shot learning has gained traction as a strategy for reducing annotation requirements in domains where labeled data are scarce. Classical approaches such as Siamese networks \cite{koch2015} and prototypical networks \cite{snell2017} learn transferable representations that generalize from a handful of examples. In agriculture, recent surveys \cite{yang2022} highlight growing interest in applying FSL to tasks such as crop classification, leaf trait extraction, and disease detection. Lagergren et al. \cite{lagergren2023} showed that few-shot methods can effectively model complex leaf venation patterns using extremely limited data. However, few-shot regression---particularly for plant growth traits---remains underexplored. Most existing efforts focus on categorical outputs or segmentation masks rather than continuous width or height estimation.

\subsection{Meta-Learning for Low-Data Adaptation}

Meta-learning aims to produce models capable of rapid adaptation to new tasks using limited support samples. The MAML framework \cite{finn2017model} has demonstrated strong performance across robotics, medical imaging, and vision tasks, and recent agricultural adaptations, such as SWE-MAML \cite{li2025}, further improve robustness in few-shot settings. Yang et al. \cite{yang2025} applied meta-learning to potato growth estimation using limited spectral data. Extensions to MAML include Meta-SGD \cite{liMetaSGD2017}, which learns per-parameter inner-loop learning rates, and MAML++ \cite{antoniou2019mamlpp}, which introduces per-step per-layer learning rates and multi-step loss for improved stability. Satrya and Yun \cite{satrya2023} showed that combining MAML with transfer learning improves regression stability across tasks with different data distributions. Despite these advances, a comparative study of MAML variants for image-based growth prediction in greenhouse environments is lacking.

\subsection{Vision Transformers and Embedding-Based Sampling}

Vision Transformers \cite{dosovitskiy2020} have been widely adopted for plant disease identification, leaf segmentation, and mobile phenotyping applications \cite{li2023,dewage2025uncertainty}. Their ability to model long-range dependencies makes them particularly suited for representing plant structure. Beyond feature extraction, ViT embeddings can support efficient sample selection. Clustering-based representative sampling, especially using soft clustering methods such as fuzzy c-means, helps capture continuous variation in plant morphology. This aligns with recent coreset selection strategies \cite{lee2024}, which emphasize preserving diversity in embedding space for object detection tasks. Recent work by Dewage et al. \cite{dewage2025representative} demonstrated that FCM-based representative sampling outperforms random selection in few-shot plant growth regression, achieving lower RMSE and more stable performance across classical machine learning models.

However, while recent work has explored representative sampling and meta-learning independently \cite{tasfe2025deep,bakr2025evaluation}, their integration within frameworks such as MAML remains underexplored, particularly across multiple plant species.
\section{Dataset Descriptions}
\label{sec:data}


\subsection{NMBU Cucumber Dataset}

\begin{figure}[t]
\centering
\includegraphics[width=0.6\columnwidth]{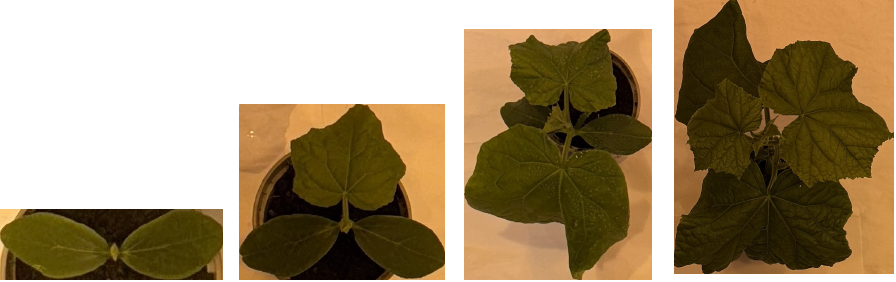}
\vspace{-2mm}
\caption{Preprocessed cucumber images from the NMBU dataset across growth stages.}
\label{fig:example_plants}
\vspace{-3mm}
\end{figure}

We introduce a novel cucumber dataset collected at the Norwegian University of Life Sciences (NMBU) under controlled greenhouse conditions. The dataset is designed to study early growth dynamics of \textit{Cucumis sativus} (var.\ Proloog). Plants were grown under regulated conditions (temperature $20 \pm 2^\circ\text{C}$, humidity $75 \pm 5\%$) with three irrigation regimes. One plant was excluded due to failed germination, resulting in 29 plants.
Images were captured daily over 16 days starting from day 8 after sowing. For each plant, a single top-view RGB image was acquired using an iPhone 16 Pro mounted $\sim$70 cm above the canopy, with a white background for consistent segmentation (Fig.~\ref{fig:example_plants}). Canopy width (mm) was measured manually, yielding $N = 464$ samples. The dataset was split 80:20 (371/93), producing 1,484 augmented training samples. The dataset will be made publicly available upon publication.

\subsection{Competition Lettuce Dataset}

\begin{figure}[t]
\centering
\includegraphics[width=0.2\columnwidth]{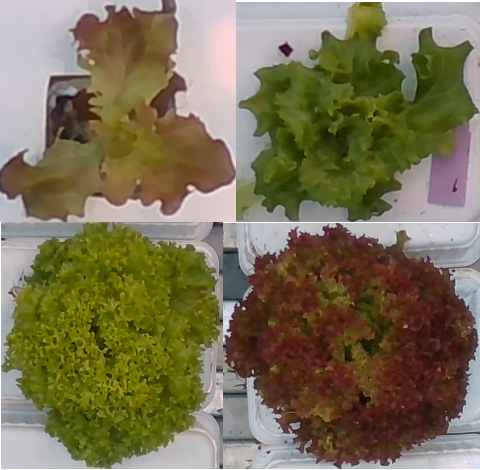}
\vspace{-2mm}
\caption{Example lettuce images showing canopy development \cite{hemming2021}.}
\label{fig:example_lettuce}
\vspace{-3mm}
\end{figure}

The lettuce dataset from the 3rd Autonomous Greenhouse Challenge \cite{petropoulou2023,hemming2021} contains $N = 388$ top-view RGB images of \textit{Lactuca sativa} (cv.\ Lugano) with canopy width annotations.
It exhibits a rosette structure distinct from cucumber (Fig.~\ref{fig:example_lettuce}). Images were split 80:20 (310/78), yielding 1,240 augmented training samples.

\section{Method}
\subsection{The Proposed Framework}
We address few-shot plant growth regression by combining representation learning, structure-aware sampling, and gradient-based meta-learning. Given a dataset of plant images $\mathcal{X} = \{x_i\}_{i=1}^N$ and corresponding continuous growth labels $\mathcal{Y} = \{y_i\}_{i=1}^N$ (e.g., canopy width in mm), the objective is to learn a model that can adapt to new samples using only a small number of labeled examples.
Figure~\ref{fig:pipeline} illustrates the overall pipeline. The key idea is to construct meaningful few-shot tasks by organizing the data in a latent morphological space, and then train a model that can rapidly adapt across these tasks.

\paragraph{Feature Representation.}
Each image $x_i$ is resized to $224 \times 224$ and passed through a pretrained Vision Transformer (ViT-Base/16) \cite{dosovitskiy2020}, used as a fixed feature extractor. This produces a compact embedding
\begin{equation}
z_i = f_{\text{ViT}}(x_i), \quad z_i \in \mathbb{R}^{768},
\end{equation}
where $z_i$ captures global morphological structure. Data augmentation (random crops, flips, rotations, and color jitter) is applied during training to improve robustness under limited data.

\paragraph{Clustering in Latent Space.}
Plant growth is inherently continuous, with gradual transitions between developmental stages. To capture this structure, we partition the embedding space $\{z_i\}_{i=1}^N$ using fuzzy c-means (FCM) clustering \cite{bezdek1981}. Unlike hard clustering, FCM assigns each sample a soft membership to every cluster. Specifically, FCM learns cluster centers $\{c_k\}_{k=1}^K$ and membership scores $u_{ik} \in [0,1]$ by minimizing
\begin{equation}
\sum_{i=1}^{N} \sum_{k=1}^{K} u_{ik}^m \|z_i - c_k\|_2^2,
\quad \text{s.t. } \sum_{k=1}^{K} u_{ik} = 1,
\end{equation}
where $m > 1$ controls the degree of softness. Here, $u_{ik}$ represents how strongly sample $i$ belongs to cluster $k$. The number of clusters $K$ is selected using a validation set to avoid test-set leakage.
This step serves two purposes: (i) it organizes samples into morphologically coherent regions, and (ii) it enables principled selection of representative examples based on membership confidence.

\paragraph{Representative Sampling and Task Construction.}
For each cluster $k$, we select a representative support set by choosing the top-$s$ samples with highest membership scores:
\begin{equation}
\mathcal{R}_k = \{(z_i, y_i) \mid i \in \text{top-}s(u_{ik})\}.
\end{equation}
Each cluster defines a task $\mathcal{T}_k$. Each task is constructed independently from a disjoint subset of the embedding space, ensuring that tasks correspond to localized regions of plant morphology. The support set $\mathcal{D}^{\text{sup}}_k = \mathcal{R}_k$ contains $s$ labeled examples, while the query set $\mathcal{D}^{\text{qry}}_k$ is sampled from the remaining points assigned to cluster $k$.

This construction enables the model to learn how growth patterns vary across different plant structures.

\paragraph{Meta-Learning.}
We train a regression model $f_\theta: \mathbb{R}^{768} \rightarrow \mathbb{R}$ that maps embeddings to growth predictions. The model is a two-layer MLP with parameters $\theta$.

Meta-learning aims to learn an initialization $\theta$ that can be quickly adapted to any task $\mathcal{T}_k$. For a given task, parameters are updated using the support set:
\begin{equation}
\theta'_k = \theta - \alpha \nabla_\theta \mathcal{L}(f_\theta, \mathcal{D}^{\text{sup}}_k),
\end{equation}
where $\alpha$ denotes the inner-loop learning rate controlling task-specific adaptation, and $\mathcal{L}$ is the mean squared error loss. The meta-objective then minimizes the query loss after adaptation:
\begin{equation}
\min_\theta \; \mathbb{E}_{\mathcal{T}_k} \left[ \mathcal{L}(f_{\theta'_k}, \mathcal{D}^{\text{qry}}_k) \right].
\end{equation}

We evaluate three variants. MAML \cite{finn2017model} uses a fixed scalar learning rate $\alpha$. Meta-SGD \cite{liMetaSGD2017} replaces $\alpha$ with learnable per-parameter rates $\boldsymbol{\lambda}$:
\begin{equation}
\theta'_k = \theta - \boldsymbol{\lambda} \odot \nabla_\theta \mathcal{L}(f_\theta, \mathcal{D}^{\text{sup}}_k),
\end{equation}
where $\boldsymbol{\lambda}$ represents learnable per-parameter step sizes.
This allows different parameters to adapt at different speeds. MAML++ \cite{antoniou2019mamlpp} further improves stability by introducing per-step learning rates and a multi-step meta-loss:
\begin{equation}
\mathcal{L}_{\text{meta}} = \sum_{j} w_j \mathcal{L}(f_{\theta_k^{(j)}}, \mathcal{D}^{\text{qry}}_k),
\end{equation}
where $\theta_k^{(j)}$ denotes parameters after $j$ inner updates and $w_j$ are weighting coefficients that balance contributions from different adaptation steps.

\paragraph{Training Protocol.}
Meta-training is performed over batches of tasks sampled from the cluster distribution. All methods use the same architecture and task construction to ensure fair comparison. Hyperparameters, including the number of clusters $K$, are selected using a validation set, and all experiments are repeated across multiple random seeds to ensure robustness.
All meta-learning variants were implemented in PyTorch using the functional API to ensure correct gradient flow through inner-loop updates. Datasets were split 80:20 (train/test, \texttt{random\_state}=42), with augmentation applied only to training data.
The Reptile baseline \cite{nichol2018} used the same MLP architecture (\texttt{hidden\_dim}=512) with tuned hyperparameters (\texttt{inner\_steps}=10, \texttt{inner\_lr}=$10^{-2}$, \texttt{meta\_lr}=0.07, \texttt{epochs}=100, \texttt{batch\_size}=8). Classical baselines—SVR \cite{cortes1995}, Random Forest \cite{breiman2001} and multilayer perceptron (MLP) \cite{rumelhart1986learning} were evaluated in both few-shot (10-shot, same support set) and full-data (grid-search-tuned) settings.
We include Reptile as a representative first-order meta-learning baseline due to its computational simplicity and widespread use. While additional baselines could be considered, our goal is to isolate the effect of gradient-based meta-learning strategies under identical task construction.
\section{Results}
\label{sec:results}

\begin{table}[t]
\centering
\footnotesize
\setlength{\tabcolsep}{3pt}
\caption{Performance comparison across methods. The top block corresponds to our proposed pipeline with different meta-learning algorithms. Best few-shot results are in bold.}
\label{tab:performance_summary}
\resizebox{\columnwidth}{!}{
\begin{tabular}{lcccc}
\toprule
& \multicolumn{2}{c}{\textbf{Cuc.}} & \multicolumn{2}{c}{\textbf{Lett.}} \\
\cmidrule(lr){2-3} \cmidrule(lr){4-5}
\textbf{Model} & RMSE & $R^2$ & RMSE & $R^2$ \\
\midrule
\multicolumn{5}{l}{\textbf{Our pipeline (meta-learning)}} \\
\textbf{MAML++ (ours)} & \textbf{35.13$\pm$1.55} & \textbf{0.783} & \textbf{29.13$\pm$1.20} & \textbf{0.773} \\
Meta-SGD (ours) & 41.29$\pm$2.82 & 0.699 & 32.54$\pm$2.40 & 0.716 \\
MAML (ours) & 42.89$\pm$1.70 & 0.676 & 29.41$\pm$1.32 & 0.769 \\
Reptile (ours) & 57.18$\pm$6.18 & 0.419 & 47.91$\pm$3.09 & 0.386 \\
\midrule
\multicolumn{5}{l}{\textit{Few-shot baselines (no meta-learning)}} \\
SVR & 74.30 & 0.030 & 64.43 & -0.107 \\
RF & 77.05$\pm$0.75 & -0.043 & 66.50$\pm$0.77 & -0.179 \\
MLP (10-shot) & 52.73$\pm$2.65 & 0.510 & 38.67$\pm$0.84 & 0.601 \\
\midrule
\multicolumn{5}{l}{\textit{Full-data baselines (upper bound)}} \\
SVR & 22.76 & 0.909 & 23.35 & 0.855 \\
RF & 31.85 & 0.822 & 23.57 & 0.852 \\
MLP & 23.04$\pm$0.95 & 0.907 & 25.25$\pm$1.64 & 0.829 \\
\bottomrule
\end{tabular}
}
\end{table}
All methods are evaluated on held-out test sets across 20 random seeds. Growth labels are standardized during training and de-standardized for evaluation, all RMSE values are reported in millimeters.
\subsection{Main Performance Comparison}
Table~\ref{tab:performance_summary} compares three settings. Our few-shot meta-learning pipeline uses ViT features, FCM-based task construction ($K=7$), and representative sampling (10 samples per cluster, 70 total), followed by MAML-based optimization. Few-shot baselines use the same setup but replace meta-learning with standard regressors (SVR, RF, MLP). Full-data baselines are trained on the entire dataset using the same ViT features, providing an upper bound.
Table~\ref{tab:performance_summary} shows that MAML++ consistently achieves the lowest RMSE among few-shot methods. On cucumber, it improves over MAML from $42.89 \pm 1.70$ mm to $35.13 \pm 1.55$ mm (18\%) and outperforms Meta-SGD by over 6 mm. On lettuce, MAML++ ($29.13 \pm 1.20$ mm) and MAML ($29.41 \pm 1.32$ mm) perform similarly.

MAML++ ranks first on both datasets, while the relative order of MAML and
Meta-SGD is dataset-dependent: Meta-SGD is second on cucumber but falls behind
MAML on lettuce. Reptile~\cite{nichol2018} performs substantially worse than
MAML on both datasets (33\% and 63\% higher RMSE on cucumber and lettuce,
respectively), highlighting the importance of second-order optimization.
MAML++ also shows the lowest variance across seeds, indicating more stable
convergence, whereas Meta-SGD is the most variable.
Classical baselines perform poorly in the 10-shot regime. Even MLP (10-shot)
underperforms (e.g., 52.73 mm on cucumber), showing that architecture alone is
insufficient without meta-learning. In contrast, full-data models (e.g., SVR:
22.76 mm, MLP: 23.04 mm) provide an upper bound. Notably, MAML++ substantially
reduces the gap to full-data performance using $<$6\% of the augmented training
set (70 labeled samples), remaining within roughly 6--13 mm of the best
full-data baseline. These trends across cucumber and lettuce indicate that the
framework generalizes across plant morphologies and imaging conditions.

\begin{figure}[t]
\centering
\includegraphics[width=\columnwidth]{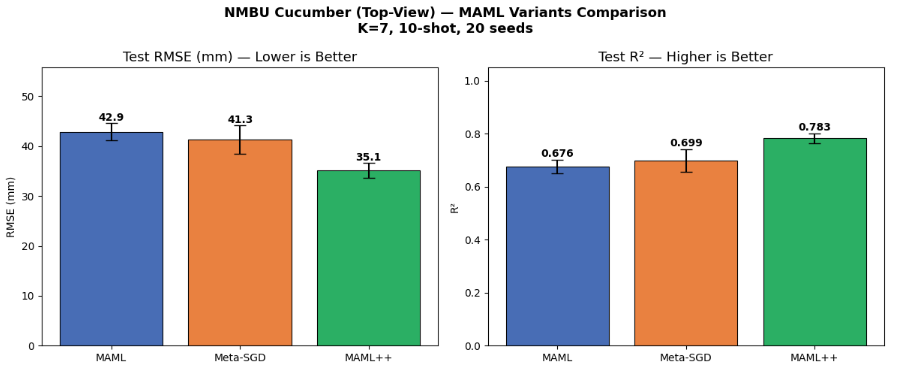}
\vspace{2pt}
\includegraphics[width=\columnwidth]{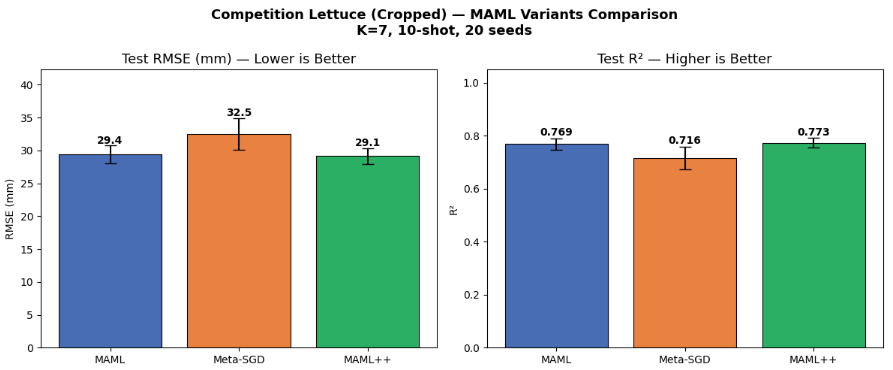}
\caption{Comparison of meta-learning variants across 20 random seeds. MAML++ achieves the lowest RMSE, with a larger margin on cucumber than lettuce.}
\label{fig:maml_variants_bar}
\end{figure}

Figure~\ref{fig:maml_variants_bar} provides a visual comparison across random seeds, confirming that MAML++ consistently achieves the lowest RMSE and exhibits lower variance, particularly on the cucumber dataset.
\subsection{Cluster Selection}

Table~\ref{tab:cluster_sweep} shows that $K=7$ yields the lowest validation RMSE on both datasets. The small difference between $K=7$ and $K=10$ indicates that performance is stable within this range and not highly sensitive to moderate changes in cluster count. This suggests that the embedding space captures plant morphology effectively, allowing consistent partitioning into a limited number of meaningful regions. In contrast, smaller values ($K=3,5$) generally lead to higher error, likely due
to insufficient representation of growth variability, although the trend is
non-monotonic on lettuce.

Cluster selection is performed exclusively on a validation split derived from the training data, avoiding test-set leakage and ensuring unbiased evaluation. This is particularly important in few-shot settings, where results are sensitive to hyperparameter tuning. Overall, validation-based selection provides a reliable and reproducible strategy, confirming that moderate cluster granularity is sufficient for effective task construction.
\begin{table}[t]
\centering
\caption{Validation-based cluster count selection over 20 random seeds.}
\label{tab:cluster_sweep}
\begin{tabular}{lccc}
\toprule
Dataset & $K$ & Val. RMSE & Selected \\
\midrule
\multirow{4}{*}{Cucumber}
& 3  & 41.541 $\pm$ 2.028 & \\
& 5  & 38.495 $\pm$ 1.767 & \\
& 7  & \textbf{36.494 $\pm$ 1.522} & \checkmark \\
& 10 & 36.816 $\pm$ 2.423 & \\
\midrule
\multirow{4}{*}{Lettuce}
& 3  & 37.151 $\pm$ 0.910 & \\
& 5  & 38.538 $\pm$ 1.864 & \\
& 7  & \textbf{34.908 $\pm$ 1.481} & \checkmark \\
& 10 & 35.040 $\pm$ 1.092 & \\
\bottomrule
\end{tabular}
\end{table}
\subsection{Ablation of Within-Cluster Support Selection Strategies}

We analyze the effect of support-set construction on few-shot performance by separating two factors: (i) how samples are distributed across clusters, and (ii) how samples are selected within each cluster.
Table~\ref{tab:balanced_unbalanced} compares balanced and unbalanced representative sampling, which control the distribution of samples across clusters. Balanced sampling enforces an equal number of samples per cluster, ensuring coverage of the embedding space, whereas unbalanced sampling selects high-membership samples globally without per-cluster constraints.
Balanced sampling consistently improves both accuracy and stability. On cucumber, RMSE decreases from 40.89 to 38.58 mm with a noticeable reduction in variance; a similar but smaller improvement is observed on lettuce (29.07 vs.\ 30.30 mm). This indicates that enforcing coverage across clusters leads to more robust task construction.

\begin{table}[t]
\centering
\caption{Balanced vs.\ unbalanced representative sampling ($K=5$). Balanced sampling enforces an equal number of samples per cluster, while unbalanced sampling selects high-membership samples globally without cluster-level constraints.}
\label{tab:balanced_unbalanced}
\begin{tabular}{lcc}
\toprule
Dataset / Setup & RMSE & Std \\
\midrule
Cucumber, balanced   & 38.576 & 1.518 \\
Cucumber, unbalanced & 40.886 & 2.298 \\
Lettuce, balanced    & 29.071 & 1.294 \\
Lettuce, unbalanced  & 30.302 & 1.087 \\
\bottomrule
\end{tabular}
\end{table}

Table~\ref{tab:representative_random} compares representative and random sampling within clusters, while keeping the number of samples per cluster fixed. Representative sampling selects the highest-membership (most prototypical) samples in each cluster, whereas random sampling selects samples uniformly within the cluster.
Differences are small on cucumber ($<1$ mm), while random sampling performs better on lettuce (27.74 vs.\ 29.41 mm), indicating no consistent advantage for representative selection.

\begin{table}[t]
\centering
\caption{Representative (highest-membership) vs.\ random support selection within clusters (fixed number of samples per cluster). This isolates the effect of intra-cluster selection independent of cluster coverage.}
\label{tab:representative_random}
\begin{tabular}{lcc}
\toprule
Dataset / Strategy & RMSE & Std \\
\midrule
Cucumber, representative & 42.886 & 1.704 \\
Cucumber, random         & 42.251 & 1.575 \\
Lettuce, representative  & 29.409 & 1.324 \\
Lettuce, random          & 27.741 & 1.176 \\
\bottomrule
\end{tabular}
\end{table}

These results suggest that support-set construction is primarily driven by cluster-level coverage, while intra-cluster selection heuristics play a secondary role compared to the meta-learning framework.

\subsection{Statistical Significance}

Table~\ref{tab:statistical_tests} reports Wilcoxon signed-rank tests across 20 random seeds. As a non-parametric test, it avoids assumptions about normality of performance differences.
MAML significantly outperforms Reptile on both datasets ($p < 0.001$, $W=0$), confirming that the large performance gap is systematic and highlights the importance of second-order optimization for regression.
The comparison between MAML and MAML++ is dataset-dependent. On cucumber, MAML++ significantly improves over MAML ($p < 0.001$), while on lettuce the difference is not significant ($p = 0.546$), indicating comparable performance within variance.
Meta-SGD does not provide consistent gains. On cucumber, its improvement over MAML is not statistically significant ($p = 0.058$), and on lettuce it performs significantly worse ($p < 0.001$), suggesting instability in learning per-parameter step sizes.
The results show that (i) second-order meta-learning consistently outperforms first-order methods, (ii) MAML++ provides additional gains in more variable settings, and (iii) Meta-SGD is not reliably beneficial in this task.

\begin{table}[t]
\centering
\caption{Wilcoxon signed-rank tests across 20 random seeds.}
\label{tab:statistical_tests}
\begin{tabular}{llcc}
\toprule
Dataset & Comparison & $W$ & $p$ \\
\midrule
\multirow{3}{*}{Cucumber}
& MAML vs.\ Reptile  & 0  & $< 0.001$ \\
& MAML vs.\ Meta-SGD & 54 & 0.058 \\
& MAML vs.\ MAML++   & 0  & $< 0.001$ \\
\midrule
\multirow{3}{*}{Lettuce}
& MAML vs.\ Reptile  & 0  & $< 0.001$ \\
& MAML vs.\ Meta-SGD & 9  & $< 0.001$ \\
& MAML vs.\ MAML++   & 88 & 0.546 \\
\bottomrule
\end{tabular}
\end{table}

\section{Discussion}
\label{sec:discussion}

The results highlight key insights for data-efficient plant growth estimation. A consistent finding is the large and statistically significant gap between second-order methods (MAML, MAML++) and first-order Reptile across both datasets. A likely explanation is that regression in this setting requires precise
parameter adaptation, which MAML captures through inner-loop differentiation,
whereas Reptile's first-order updates are insufficient. This indicates that second-order methods are more effective for few-shot regression.

Although MAML++ performs best, its advantage over MAML is dataset-dependent. Gains are substantial on cucumber but negligible on lettuce, which may relate to differences in dataset variability.
Moreover, balanced sampling improves stability by ensuring coverage across the embedding space, while representative (FCM-based) sampling does not consistently outperform random selection, suggesting diversity is more important than prototypicality.
Additionally, meta-learning substantially reduces label requirements. MAML++ achieves competitive performance using less than 6\% of the augmented training data, which is valuable in greenhouse settings with limited annotations.
Consistent trends across cucumber and lettuce indicate generalization across plant structures and imaging conditions.
\section{Conclusion}

We presented a data-efficient framework for plant growth estimation that combines ViT-based representations, FCM-based task construction, and gradient-based meta-learning. Across two plant datasets with distinct morphology and imaging conditions, MAML++ achieved the best overall few-shot performance, reaching 35.13 mm RMSE on cucumber and 29.13 mm on lettuce.
The experiments show that second-order meta-learning is effective for low-data plant growth regression, while the benefit of structured representative sampling is limited and dataset-dependent. The validation-based cluster-selection protocol further ensures that task construction is performed without test-set leakage, improving the reliability of model selection.
This study is limited to controlled greenhouse imagery and single-trait regression. Future work should evaluate broader crop diversity, temporal modeling, and uncertainty-aware prediction to support deployment in real greenhouse monitoring systems.

\bibliographystyle{IEEEtran}
\bibliography{IEEEabrv, bibliography}{}

\end{document}